\documentclass[final, runningheads]{llncs}

\usepackage[T1]{fontenc}
\usepackage{graphicx}
\usepackage{color}

\usepackage{amsmath, amssymb, mathtools}
\usepackage[british]{babel}
\usepackage[english=british]{csquotes}
\MakeOuterQuote{"}
\usepackage[draft=false,hidelinks]{hyperref} 

\usepackage{tikz}
\usetikzlibrary{calc, patterns, arrows, shapes, positioning, fit, cd, decorations.pathreplacing,calligraphy}
\tikzset{every picture/.style={thick}}
\usepackage{tikzpeople}

\usepackage{acronym}
\usepackage{xspace}
\usepackage[nameinlink, capitalise]{cleveref}
\usepackage{fixme} 
\usepackage[final]{microtype}
\usepackage{varwidth}
\usepackage{lexref}
\usetikzlibrary{calc, arrows, patterns, shapes, positioning, arrows.meta,fit}

\spnewtheorem*{example}{Example}{\bfseries}{\rmfamily}

\usepackage{version}
\excludeversion{fullversion}
\includeversion{shortversion}

\acrodef{AI}{artificial intelligence}
\acrodef{WSM}{Wall Street Market}
\acrodef{USPS}{US Postal Service}
\acrodef{BPPC}{bitcoin payment processing company}
\acrodef{KYC}{Know-Your-Customer}
\acrodef{CQ}{critical question}
\acrodef{GCCP}{German Code of Criminal Procedure}
\DeclareLex{GCCP}{\ac{GCCP}}[\acl{GCCP}]
\RenewLexShortcut{DispPrefixMain}{Section}[]
\RenewLexShortcut{DispsPrefixMain}{Sections}[]

\usepackage{etoolbox} 
\makeatletter
\patchcmd{\ps@headings}{\rlap{\thepage}\hspace{\headlineindent}}{}{}{}
\patchcmd{\ps@headings}{\hspace{\headlineindent}\llap{\thepage}}{}{}{}
\makeatother

\crefname{cq}{{\acs{CQ}}}{{\acp{CQ}}}
\crefname{scheme}{Scheme}{Schemes}
\usepackage[font=small]{caption}
\DeclareCaptionType[placement={h!},within=none]{scheme}[Scheme][List of Schemes]

\usepackage{booktabs}

\newcommand{\schemeB}[3]{%
  \begin{tabular}{ll}
    Premise: & #1 \\
    Premise: & #2 \\
    \midrule
    Conclusion: & #3\\
  \end{tabular}
}

\newcommand{\schemeT}[4]{%
  \begin{tabular}{ll}
    Premise: & #1 \\
    Premise: & #2 \\
    Premise: & #3 \\
    \midrule
    Conclusion: & #4\\
  \end{tabular}
}
\usepackage{placeins}

\newcommand\btc{Bitcoin\xspace}
\newcommand\ie{i.e.\xspace}

\newcommand\pk{\ensuremath{\mathsf{pk}}}

\usepackage[backend=biber,style=lncs,maxbibnames=3]{biblatex}
\bibliography{waltonchain}

\fxuselayouts{inline,nomargin}
\fxusetheme{color}
\FXRegisterAuthor{ns}{ans}{Nici}
\FXRegisterAuthor{mg}{amg}{\color{red}Merlin}
\FXRegisterAuthor{vr}{avr}{\color{blue}Vicky}
\FXRegisterAuthor{dd}{add}{\color{magenta}Dominic}
\FXRegisterAuthor{jg}{ajg}{\color{cyan}Jan}

\title{Argumentation Schemes for Blockchain Deanonymization}
\author{Dominic~Deuber\orcidID{0000-0002-8177-0562} \and Jan~Gruber\orcidID{0000-0003-1862-2900} \and Merlin~Humml\orcidID{0000-0002-2251-8519} \and Viktoria~Ronge \and Nicole~Scheler}
\authorrunning{D.\ Deuber et al.}

\institute{Friedrich-Alexander-Universität Erlangen-Nürnberg, Erlangen, Germany\\
\email{\{firstname.lastname\}@fau.de}}

\begin{document}

\maketitle

\begin{abstract}
  Cryptocurrency forensics became standard tools for law enforcement.
  Their basic idea is to deanonymise cryptocurrency transactions to identify the people behind them.
  Cryptocurrency deanonymisation techniques are often based on premises that largely remain implicit, especially in legal practice.
  On the one hand, this implicitness complicates investigations. On the other hand, it can have far-reaching consequences for the rights of those affected.
Argumentation schemes could remedy this untenable situation by rendering underlying premises transparent. Additionally, they can aid in critically evaluating the probative value of any results obtained by cryptocurrency deanonymisation techniques.
In the argumentation theory and \acs{AI} community, argumentation schemes are influential as they state implicit premises for different types of arguments.
  Through their \acl*{CQ}s, they aid the argumentation participants in critically evaluating arguments.
We specialise the notion of argumentation schemes to legal reasoning about cryptocurrency deanonymisation. 
Furthermore, we demonstrate the applicability of the resulting schemes through an exemplary real-world case.
Ultimately, we envision that using our schemes in legal practice can solidify the evidential value of blockchain investigations as well as uncover and help address uncertainty in underlying premises -- thus contributing to protect the rights of those affected by cryptocurrency forensics.
\keywords{Argumentation \and Legal Reasoning \and Blockchain Analysis.}
\end{abstract}

\section{Introduction}\label{sec:introduction}
``Follow the money'' is arguably the central investigation strategy for any profit-driven offence~\cite{wechsler2001}.
Analysing flows of incriminated money is crucial to understand the business
models and inner workings of organised crime groups, the hierarchy of the
involved entities, and finally, identifying the groups' members.
However, the fight against money laundering is challenging, and criminals utilising 
virtual currencies%
\begin{fullversion}
, also referred to as cryptocurrencies,
\end{fullversion}
as early adopters aggravate the situation even further.
\begin{fullversion}
Nowadays Bitcoin and the consortia of ``alt-coins'' have to be considered key facilitators
of cybercrime for years~\cite{iocta2017}. But not only the cyber-dimension of
crime profits enormously from the utilisation of virtual currencies, increasing
news reports on traditional \acp{OCG}, like Albanian gangs, the Nigerian Black Axe or
South-American cartels utilising virtual currencies to maximise their profits
and minimise the risk of
discovery~\cite{times2019,reuters2020,bbc2021}.
\end{fullversion}
While law enforcement agencies need to expend many resources to follow complex transnational flows of fiat currencies,
blockchain-based investigations impose even further challenges.
These challenges arise from the fact that cryptocurrencies are generally pseudonymous, with some even being anonymous. 
Bitcoin~\cite{nakamoto2008bitcoin} is arguably the most famous and widespread cryptocurrency~--~both for lawful economic purposes and criminal activities~\cite{Europol2021}.
Already in the early days of Bitcoin, it was shown that the currency is not anonymous because it is possible to link multiple pseudonyms belonging to the same person~\cite{FC:AKRSC13,meiklejohn2013fistful,reid2013analysis}.
However, also supposedly anonymous cryptocurrencies, such as Monero~\cite{monero} or Zcash~\cite{zcash}, have been target of deanonymisation attacks~\cite{USENIX:KYMM18,PoPETS:MSHLHSHHMNC18}. 
What all attacks on Bitcoin, Monero, and Zcash have in common is that
they are based on partly unreliable assumptions~\cite{DBLP:journals/popets/DeuberRR22}.
The reliability of these assumptions determines the quality of the
results of an attack. In legal practice, those assumptions are
critical for inferring the evidential value of the deanonymisation of
a perpetrator. However, no standard practice for deriving and
discussing the reliability of those analysis results has been proposed
yet.
Therefore, we propose argumentation schemes for assessing the reliability of investigations on the \btc{} blockchain -- thus bridging practical cryptocurrency forensics and its scientific analysis. 


\subsection{Related Work}\label{sec:related-work} 
\emph{Argumentation schemes}~\cite{walton_reed_macagno_2008} as a way to classify arguments by their underlying principles of convincingness have been influential in the argumentation theory and the \acl{AI} community~\cite{DBLP:journals/argcom/Macagno21}.
They present the various types of arguments as informal deduction rules together with accompanying \emph{critical questions} to aid a human reasoner in evaluating arguments of the respective type. 

Given that expert testimonies, as well as the court process itself, is a form of argumentation, it is not surprising that argumentation schemes were applied to legal processes~\cite{DBLP:journals/argcom/AtkinsonB21}.
\textcite{waltonlegal2018} gives a detailed overview of the applicability of many argumentation schemes to representing and analysing legal processes.
Apart from the argumentation schemes, there are other informal argument schemes like the ones proposed by \textcite{Wagemans_2016}; however, they focus more on the classification of arguments rather than human comprehension.
There have also been more formal -- and even automated -- approaches to legal reasoning based on argumentation theory \cite{DBLP:journals/flap/Prakken17,DBLP:journals/argcom/AtkinsonB21}.
However, our goal is not to automate parts of the legal process but to aid in evaluating statements about blockchain deanonymisation.
While software automates blockchain deanonymisation (e.g.\ Chainalysis Reactor~\cite{chainalysis}), in the end, legal decision makers, i.e.\ humans, need to evaluate the reliability of the obtained findings.

Postulating application-tailored argumentation schemes to capture specialised forms of argument is common practice.
\textcite{Parsons_2014} introduce schemes to reason about trust in entities to specialise arguments building on statements.
Another example from the medical field is specific argumentation schemes to reason about treatment choices in order to aid doctors in their decision making and producing automated patient specific recommendations~\cite{sanchez2019argumentation,DBLP:journals/argcom/SassoonKMP21}.

On the legal side, the evidence must be critically evaluated as investigative measures justified by unreliable results potentially impinge upon the fundamental rights of the suspects~\cite{DBLP:journals/cybersecurity/Rueckert19}.
\textcite{froewis2019cojo} provide key requirements that must be satisfied to safeguard the evidential value of cryptocurrency investigations; one of them being reliability. They suggest specific measures to achieve reliability, such as sharing any information necessary to assess reliability, without discussing how they can be implemented in practice.
As a step in that direction, \textcite{DBLP:journals/popets/DeuberRR22} provide a taxonomy for the different assumptions underlying deanonymisation attacks on cryptocurrency users -- while only briefly discussing their taxonomy's applicability in legal practice.

\subsection{Contribution}\label{sec:our-approach}
In legal practice, the lack of a profound framework means that there is no standard way to reason about the reliability of findings from blockchain-based investigations.
Less reliable findings might entail two issues:
First, results with low reliability might not establish the degree of suspicion required by subsequent investigative measures and thus render them unlawful.
In the worst case, any evidence obtained from unlawful investigations might be inadmissible in court~--~depending on the exclusionary rules of the respective jurisdictions.
Second, even if evidence might be admissible, low reliability corresponds to low evidential value, and thus the evidence might not be sufficient for a conviction.
Given that any findings and the blockchain investigation itself are highly abstract for most parties involved, there needs to be a common ground between technical analysts, investigators, and other legal practitioners to assess these findings.


Our contribution is the application of tailored argumentation
schemes to assess heuristics employed in investigations based on the \btc{} blockchain to deanonymise criminal users.
The schemes render the taxonomy proposed by~\textcite{DBLP:journals/popets/DeuberRR22} broadly accessible and easy to use in practice.
By presenting the implicit and explicit premises of those heuristics, our argumentation schemes enable all parties involved in the legal process to assess evidential value systematically.
Thus, the schemes can potentially render blockchain-based analyses of \btc{} transactions more comprehensible and the findings more reliable and conclusive.
\section{Preliminaries}\label{sec:background}


\subsection{\btc (BTC)}\label{sec:btc}
Bitcoin \cite{nakamoto2008bitcoin} is a cryptocurrency.
At its core are transactions that, in their most basic form, are payments. 
In contrast to fiat currencies, Bitcoin employs a decentralised ledger of transactions. 
Decentralised means that there is no central authority issuing new units of the currency or settling transactions.
Instead, parties maintain the ledger in a peer-to-peer network~--~a network where all parties are clients and servers simultaneously.
The transactions are organised in blocks
, which is why the ledger is also referred to as a  blockchain.
Using a consensus mechanism, the network agrees on which blocks, \ie particularly transactions, should extend the ledger.
The network nodes participating in this consensus mechanism are called \emph{miners}.

\begin{figure}[h]
\centering
\begin{tikzpicture}
[every node/.style={minimum height=0.7cm, minimum width=2.2cm}, transform shape]

\node (tx) {TX};

\node [anchor=west, yshift=-15](helpR) at($(tx.south)+(.2cm,0)$){\phantom{$tx_{hash}$, $out_{id}$}};
\node [anchor=east, yshift=-15](helpL) at($(tx.south)-(.2cm,0)$){\phantom{$tx_{hash}$, $out_{id}$}};

\node (inpL) at (helpL){In};
\node [below=of inpL, draw, yshift=30](inp) {$tx_{hash}$, $out_{id}$};

\node (outL)at(helpR) {Out};
\node [below=of outL, draw, yshift=30](out1) {$h_{pk_1}$, $v_1$\,{BTC}};
\node [below=of out1, draw, yshift=20](out2) {$h_{pk_2}$, $v_2$\,{BTC}};

\coordinate[left=of inp](arrowL);
\draw[-Latex] (arrowL) -- (inp);

\coordinate[right=of out1](arrowR);
\draw[-Latex] (out1) -- (arrowR);

\coordinate[right=of out2](arrowR);
\draw[-Latex] (out2) -- (arrowR);

\node[inner sep=.2cm, fit=(inpL)(inp)(outL)(out1)(out2)(helpR)(helpL), draw](fit) {};

\draw[dashed] (fit.south) edge (fit.north);
\end{tikzpicture}

\caption{\btc transaction}\label{fig:tx_btc}
\end{figure}

\paragraph{Transactions}\label{subsec:txs} consist of a list of inputs and outputs.
An output usually states an amount of \btc ($v$\,BTC) and the hash $h_{\pk}$ of a public key $\pk$, which is also referred to as address $a$.
The public key is part of a digital signature scheme. 
Such schemes use public and secret key pairs~--~anyone can check the validity of a signature with respect to some public key, while only the one knowing the corresponding secret key can create a valid signature.
An input is a reference to an output of another transaction, which is uniquely described by the hash  $tx_{hash}$ of that other transaction and the position $out_{id}$ of the output in the transaction's list of outputs.
An example of a transaction with one input and two outputs is given in~\cref{fig:tx_btc}.
Usually, transactions have several in- and outputs.
Spending the first output of this transaction with an amount of $v_1$ \btc requires providing a public key $\pk'$ whose hash equals $h_{\pk_1}$ and a signature that verifies under $\pk'$.
This mechanism ensures that, in general, there are no unauthorized transactions, as knowledge of the corresponding secret keys is required to issue a transaction.
A property of Bitcoin is that the input amount of a transaction is always consumed entirely.
Thus, the second output of the transaction might be a so-called \emph{change} output.
A change output pays back to the sender(s) the difference between its input amounts and the amount that the recipient(s) should receive.

\paragraph{Wallets}
in \btc can be seen as a collection of several addresses which belong to the same entity.
On a technical level, a wallet is often referred to as software that generates and stores the private keys corresponding to different addresses and allows creating new addresses and issuing transactions.
By only inspecting transactions on the blockchain, it is not immediately obvious which addresses belong to the same wallet.

\paragraph{CoinJoin} transactions are a special type of transaction that tries to add anonymity to Bitcoin.
The idea is to combine inputs from multiple entities while at the same time having equally valued outputs~\cite{maxwell2013coinjoin}. In \btc{}, the concept of having transactions with inputs from multiple users to hinder linking is called \emph{mixing}.


\subsection{\btc{} Investigations}
Research has shown early on that \btc is not anonymous but pseudonymous, as it is possible to cluster addresses that are likely to be controlled by the same entity, referred to as \emph{address clustering}.
The most important address-clustering heuristics are the \emph{multi-input heuristic}~\cite{FC:AKRSC13,meiklejohn2013fistful,reid2013analysis} and the \emph{change-address heuristic}~\cite{meiklejohn2013fistful,FC:AKRSC13,USENIX:KYMM18}. 
The multi-input heuristic states that all inputs of a transaction are controlled by the same entity -- as already mentioned in Bitcoin's whitepaper~\cite{nakamoto2008bitcoin}.
The multi-input heuristic should not be applied to CoinJoin transactions as they are issued by multiple entities by design.
The change-address heuristics utilise that change often occurs in \btc (see \cref{subsec:txs}).

The main objective of blockchain investigations is \emph{re-identification}, that is to determine the natural or legal person who controls an address cluster.
This is especially relevant for law enforcement trying to identify persons connected to flows of incriminated virtual currencies. By tracing such transactions and conducting address clustering, they might identify a single relevant address cluster.
As addresses typically do not contain any personally identifiable information, the investigation requires re-identification. To facilitate re-identification, address clusters are usually connected with off-chain information -- a process also referred to as \emph{attribution tagging}~\cite{froewis2019cojo}.
As its name implies, the tagged information in attribution tagging can be used to identify the actual entity. 
In practice, the arguably most important attribution information is that an address cluster is related to some cryptocurrency \textit{exchange} -- a platform to exchange, buy or sell cryptocurrencies -- as law enforcement might request the respective customer data from this exchange. 

\subsection{Legal Background}\label{sec:legal}
Many states committed themselves to the fight against cybercrime by ratifying the Convention on Cybercrime~\cite{conventiononcybercrime2001}.
This commitment includes establishing cybercrime offences under domestic law as well as providing investigative measures to enable the prosecution of such offences -- while simultaneously protecting fundamental human rights and liberties.
The actual balance between the interests of law enforcement and human rights is dictated by the domestic laws of the ratifying states.
However, the legal issues discussed in this section are not specific to a particular jurisdiction or legal system. This is illustrated by using the US as an example of a common-law jurisdiction and Germany as an example of a civil-law jurisdiction; both states have ratified the convention.
The starting point for our discussion is the following example case of a typical blockchain-based investigation:
\begin{example}\label{exp:case}
  Investigators seized a darknet marketplace and recovered a local Bitcoin wallet that was presumably used to pay the marketplace's operator.
  The investigators then used blockchain analysis to discover the wallet which was used by the operator to receive payments.
  While the discovered operator wallet is a local wallet, the operator is suspected of using another wallet at a cryptocurrency exchange to convert Bitcoin into fiat currency.
To prevent that the exchange wallet can be linked to the incriminated local wallet, the operator mixed the funds prior to the transfer.
Through blockchain analysis, the investigators nevertheless managed to establish a link between the incriminated local wallet and the exchange wallet.
Next, the investigators issued a request for the disclosure of customer data to the exchange -- which collected them as part of their employed \acl{KYC} policy to comply with anti-money-laundering laws. 
The goal of this request was to find the natural person that controls the incriminated local wallet.
  After having identified this suspected operator, the investigators conducted electronic surveillance and executed a search of the suspect's premises.
\end{example}

In summary, the investigative measures used in the example were the blockchain analysis, a request for the disclosure of customer data, electronic surveillance, and a search of premises.
In general, such investigative measures have in common that they require a specific degree of suspicion in order to protect the rights of the targeted person.

Under German law, an \emph{initial suspicion} is sufficient to justify a blockchain analysis (according to~\GCCP{161, 163}
, \cite{Saff/Rueck/TelkobeiBitcoins,Grzywotz/Koehler/Rueckert}) or a request for the disclosure of customer data (according to~\GCCP{100j}).
An initial suspicion must be based on a conclusive and established factual basis (factual quality).
Due to lax requirements, these measures may be directed not only against the suspected person but also against other third parties that might be somehow connected~\cite{MueKo152,MueKo100j}.
There are stronger requirements regarding electronic surveillance pursuant to~\GCCP{100a} or a search of premises pursuant to~\GCCP{102}.
Beyond the mere \textquote{possibility} of the commission of a crime, in these cases, the suspicion of the crime must be specific and individualised (so-called \emph{qualified initial suspicion}) as well as \textquote{probable}~\cite{MueKo100a, MueKo102}.
These measures have to be directed only against the accused person \cite{MueKo100a} and may only involve other persons who are directly connected to the accused person or involved in the crime~(see~Sections~\npGCCPns{100a}~(3)~and~\npGCCP{103}). 

Under US law, especially the requirements for the analysis of blockchain data and a request for the disclosure of customer data differ significantly from German law.
However, this does not affect the legal issues raised by blockchain analyses, as we will point out below.
Both blockchain analyses and the request for the disclosure of customer data are not subject to the probable cause requirement of the Fourth Amendment, given that the \emph{third-party doctrine} applies~\cite{case2020gratkowski}.
However, electronic surveillance and search of premises are subject to the Fourth Amendment and therefore require \emph{probable cause} as the degree of suspicion.
The Fourth Amendment demands the suspicion to be particularised with respect to the person under surveillance, being searched, or specific things to be seized.


The most important legal issue concerning blockchain analysis in practice is whether or not the findings of the analysis can establish the required degree of suspicion for subsequent investigative measures.
Therefore, the lower requirements for blockchain analysis or a request for the disclosure of customer data under US law do not matter, as at least subsequent measures -- such as searches of premises -- require similar degrees of suspicion as under German law.
Thus, the only difference under US law is that the legal issue arises later in the investigation.

To illustrate the legal issue, we return to the example of the darknet marketplace operator.
Here, a blockchain analysis was used to link an incriminated wallet to an exchange service.
Next, disclosure of customer data was requested from the exchange.
Imagine that solely based on the linkage of the wallets, further investigative measures are conducted against the natural person identified by the customer data.
If those measures are electronic surveillance or searches of premises, the required suspicion must be particularised against the person targeted by the measures, both under German and US law.
If it is unreliable, blockchain analysis might fail to establish this particularised suspicion.
Imagine that the analysis is based on the multi-input heuristic, but the heuristic is applied to CoinJoin transactions. In this case, the analysis would definitely yield false positives as CoinJoin transactions are issued by multiple entities by design.
False positives might render the individualisation insufficient and thus the respective investigative measure unlawful.

To summarise, certain invasive and targeted investigative measures require a degree of suspicion that is individualised with respect to the target of these measures.
Blockchain analysis based on uncertain assumptions might lead to unreliable findings that are not sufficient to establish the individualisation and thus the required degree of suspicion for subsequent investigative measures.
If investigative measures are conducted without the necessary degree of suspicion, they are unlawful and thus might render obtained evidence inadmissible -- depending on the exclusionary rules of the respective jurisdiction.

\subsection{Argumentation Schemes}\label{sec:argum-schemes}
Argumentation schemes classify arguments by their warrant in the sense of \textcite{Toulmin58} -- i.e.\ by their principle of convincingness.
They are presented as informal presumptive deduction rules inferring plausible truth of a conclusion from truth of multiple premises~\cite{walton_reed_macagno_2008}.
For example, the \emph{\nameref{fig:arg-abductive}} is tailored towards reconstructing the cause \(E\) for a set \(F\) of observed findings.

\begin{scheme}
   \schemeT{$F$ is a finding or given set of facts.}{$E$ is a satisfactory explanation of $F$.}{No alternative explanation $E'$ given so far is as satisfactory as $E$.}{Therefore, $E$ is plausible as hypothesis.}
  \caption[Argument from Abductive Inference]{Argument from Abductive Inference~\cite{walton_reed_macagno_2008}}
  \label{fig:arg-abductive}
\end{scheme}
\noindent In addition to the deduction rule representing the informal shape of the argument, an argumentation scheme specifies \emph{\acfp{CQ}} as ways to attack an argument based on the scheme.
The critical questions aid both the producer and the receiver of arguments by suggesting relevant statements to present or ask about.
There are usually \aclp{CQ} attacking the individual premises or the conclusion of the argument, as well as ones attacking the applicability of the scheme.
Consider for example the \acp{CQ} of the~\nameref{fig:arg-abductive}:

\begin{scheme}\ContinuedFloat
\begin{enumerate}
\item \label[cq]{enum:arg-abductive-1} How satisfactory is $E$ as an explanation of $F$, apart from the alternative explanations available so far in the dialogue?
\item \label[cq]{enum:arg-abductive-2} How much better an explanation is $E$ than the alternative explanations available so far in the dialogue?
\item \label[cq]{enum:arg-abductive-3} How far has the dialogue progressed?
  If the dialogue is an inquiry, how thorough has the investigation of the case been?
\item \label[cq]{enum:arg-abductive-4} Would it be better to continue the dialogue further, instead of drawing a conclusion at this point?
\end{enumerate}
\caption{\Aclp{CQ} of Argument from Abductive Inference}
\end{scheme}
\noindent \Cref{enum:arg-abductive-1,enum:arg-abductive-2} are direct attacks on truth of premises of the rule.
\Cref{enum:arg-abductive-3,enum:arg-abductive-4} are specific attacks based on the idea that there could be other explanations not yet put forth due to the temporal nature of argumentative dialogues.

By making premises and possible flaws of an argument explicit, argumentation schemes aid critical discussion of expert statements by legal decision-makers and other practitioners without the need for deep understanding of the underlying topic.
For judging the reliability of a claim from blockchain analysis, it is particularly helpful to have transparency with regards to the underlying assumptions as they have to be judged on a case-by-case basis~\cite{DBLP:journals/popets/DeuberRR22}.
This added transparency can also increase the evidential value of such findings if the reliability of dependent information is sufficiently well established.

\section{Our Argumentation Schemes}\label{sec:our-argum-schemes}

In criminal investigations, blockchain analyses are typically conducted to establish a link between an entity and a criminal offence through involved cryptocurrency addresses.
As stated in~\cref{sec:related-work}, there exists software that could establish such links in an automated manner.
However, the methods used by it, as well as the employed heuristics, remain regularly opaque. Such insufficient traceability is contrary to the requirements of legal proceedings, which require a high degree of explainability and intelligibility.
For this purpose, we present a custom argumentation scheme to argue the involvement of an entity in an offence from the control of an address that is connected to that offence (see~\cref{fig:suspicion}).

\begin{scheme}
  \schemeB{Address \(A\) is connected to offence \(O\)}{Entity \(E\) controls address \(A\)}{Entity \(E\) is connected to offence \(O\)}
  \begin{enumerate}
  \item \label[cq]{enum:arg-suspicion:q1} Which circumstantial evidence indicates that entity \(E\) controls address \(A\)?
  \item \label[cq]{enum:arg-suspicion:q2} Could it be that at the time of offence \(O\) someone else controlled address \(A\) instead of entity \(E\)?
  \item \label[cq]{enum:arg-suspicion:q3} How was address \(A\) connected to offence \(O\) that \(E\)'s involvement is indicated?
  \item \label[cq]{enum:arg-suspicion:q4} Are there other indicators that \(E\) is connect to offence \(O\)?
  \end{enumerate}
  \caption{Suspicion through Address Control}

  \label{fig:suspicion}
\end{scheme}
\noindent We do not need a custom argumentation scheme to represent linking an entity with an address by requesting data from a cryptocurrency exchange, as this is covered by \emph{Argument from Position to Know}~\cite{walton_reed_macagno_2008}.
This standard scheme covers this case, as exchanges typically collect the personal information their customers' personal information as part of \acl{KYC} policies and are thereby in a position to know who the customer using an account is.

To establish a link between addresses, there are software tools implementing various heuristics, such as the multi-input heuristic or change heuristics, which are arguably used by investigators~\cite{DBLP:journals/popets/DeuberRR22}.
We pose the \emph{\nameref{fig:blackbox}} scheme to represent arguments based on such a software tool to establish the link between addresses and thereby forming clusters.

\begin{scheme}
  \schemeT{Software \(S\) establishes a link between address \(A_1\) and address \(A_2\)}{Software \(S\) is reliable}{Entity \(E\) controls address \(A_1\)}{Entity \(E\) controls address \(A_2\)}
  \begin{enumerate}
  \item \label[cq]{enum:arg-blackbox:q1} How does software \(S\) establish the link?
  \item \label[cq]{enum:arg-blackbox:q2} How reliable is software \(S\)? Why is software \(S\) considered reliable?
  \item \label[cq]{enum:arg-blackbox:q3} Could this link be also established without the use of software \(S\), e.g. by using a different software, human-reasoning with the multi-input heuristic, or other non-blackbox methods?
  \item \label[cq]{enum:arg-blackbox:q4} What evidence exists for entity \(E\) controlling \(A_1\)?
  \item \label[cq]{enum:arg-blackbox:q5} Are there other indicators that \(E\) might control \(A_2\)?
  \end{enumerate}
  \caption{Cluster from Software}
  \label{fig:blackbox}
\end{scheme}
%
%
\noindent Naturally, it is not enough for a software tool to establish a link between addresses without further explanations and evidence backing that claim.
Analysts face a myriad of transactions when conducting blockchain analyses.
They must assess the results presented by the software for criminalistic and legal reasons.
First, analysts must understand the software's processes to infer investigative leads, find connections, and form hypotheses~--~tasks that cannot be entirely automated.
Second, only when understanding the software's results can analysts apply their knowledge of criminal tactics eventually employed by perpetrators, question the results, and falsify hypotheses they previously posed.
Finally, from a legal perspective, the rightfulness of the analysis is crucial, as it affects the lawfulness of further investigations in the pre-trial stages and the evidential value of obtained findings in the actual trial~\cite{DBLP:journals/popets/DeuberRR22}.
However, assessing the results would require that the employed deanonymization software discloses the assumptions relied on in the analysis~--~which is typically not done at all.
Therefore, an investigator would back the findings of the software by manual analysis in case the software does not disclose the reasons for linking addresses.
To represent the claims from manual analysis, we present two exemplary schemes that capture the use of the multi-input (see~\cref{fig:multi-input}) and the change-address heuristic (see~\cref{fig:change-address}), respectively.

\begin{scheme}
  \schemeB{Transaction \(T\) has multiple input addresses}{Entity \(E\) controls some input addresses of \(T\)}{Entity \(E\) controls all input addresses of \(T\)}
  \begin{enumerate}
  \item \label[cq]{enum:arg-mih:q1}Could \(T\) be a CoinJoin transaction?
  \item Could it be that another entity \(F\) shares secret keys with \(E\) and thereby can control other or all inputs of \(T\)?
  \item Which input addresses of transaction \(T\) does entity \(E\) control? What evidence is there for \(E\) controlling these addresses?
  \item Are there other indicators that \(E\) might control other input addresses of \(T\)?
  \end{enumerate}
  \caption{Cluster from Multi-Input}
  \label{fig:multi-input}
\end{scheme}
\begin{scheme}
  \schemeT{Transaction \(T\) has multiple output addresses}{Output address \(C\) is a \emph{change} address of transaction \(T\)}{Entity \(E\) controls all input addresses of \(T\)}{Entity \(E\) also controls \emph{change} address \(C\)}
  \begin{enumerate}
  \item Could \(T\) just have multiple distinct benefactors? Could the change for example be donated to a supported unrelated entity?
  \item What evidence is there suggesting that client software was used which generates a fresh change address for every new transaction?
  \item Are there other indicators that \(E\) controls address \(C\)?
  \end{enumerate}
  \caption{Cluster by Change-Address}
  \label{fig:change-address}
\end{scheme}
\noindent For brevity, the argumentation schemes presented in this section only cover the most common \btc{} blockchain analysis heuristics used in practice and especially do not cover non-blockchain-specific reasoning.
For the latter, we can use the vast array of pre-existing schemes~\cite{walton_reed_macagno_2008}.
Together, these schemes can be applied to represent reasoning about \btc{} blockchain investigations in practice, as we will show in~\cref{sec:appl-wsm}.
%
\section{Application in the \acl{WSM} Case}\label{sec:appl-wsm}
In order to illustrate our approach and its practical implications
, we present
the argumentation behind the investigative results of the proceedings against
one of the administrators of the infamous \acf{WSM}.
\Ac{WSM} was one of the largest darknet marketplaces on which illegal narcotics,
financial data, hacking software as well as counterfeit goods were traded between
approximately 2016 and its seizure in 2019~\cite{doj2019wsm}. Besides technical
surveillance measures, blockchain-based investigations of Bitcoin transactions conducted by
the \ac{USPS} were decisive in identifying the administrators
operating the marketplace~\cite{complaint2019wsm}.

\begin{figure}[ht]
  \centering
  \begin{tikzpicture}[every node/.style = {font=\scriptsize}, imgnode/.style={inner sep=0pt, align={center}}, every edge/.style = {line width=.4pt,draw=black!85, solid,shorten <= 2mm, shorten >= 2mm}, scale=1, transform shape]
    \coordinate (hansacoord) at (-8,5);
    \coordinate (dudebuycoord) at ($(hansacoord) + (0,2)$);
    \coordinate (w2coord) at ($(hansacoord)!.5!(dudebuycoord) + (2,0)$);
    \coordinate (w1coord) at ($(w2coord) + (1.7,0.75)$);
    \coordinate (w4coord) at ($(w2coord) + (1.7,-0.75)$);
    \coordinate (mixercoord) at ($(w2coord) + (3.6,0)$);
    \coordinate (bpcccoord) at ($(mixercoord) + (1.5,1)$);
    \coordinate (emailcoord) at ($(bpcccoord) + (3,0)$);
    \coordinate (shoppingcoord) at ($(bpcccoord)!.5!(emailcoord) + (0,-1)$);

    \coordinate (theonecoord) at ($(dudebuycoord) + (0,-2)$);
    \coordinate (wsmcoord) at ($(theonecoord) + (2,-1)$);
    \coordinate (w5coord) at ($(wsmcoord) + (4,0)$);
    \coordinate (pgpcoord) at ($(hansacoord)!.5!(wsmcoord)$);

    \node[imgnode, label=below:{Hansa}] (hansa) at (hansacoord) {\includegraphics[width=1cm]{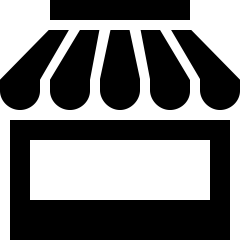}};
    \node[imgnode, label=below:{dudebuy}] (dudebuy) at (dudebuycoord) {\includegraphics[width=0.75cm]{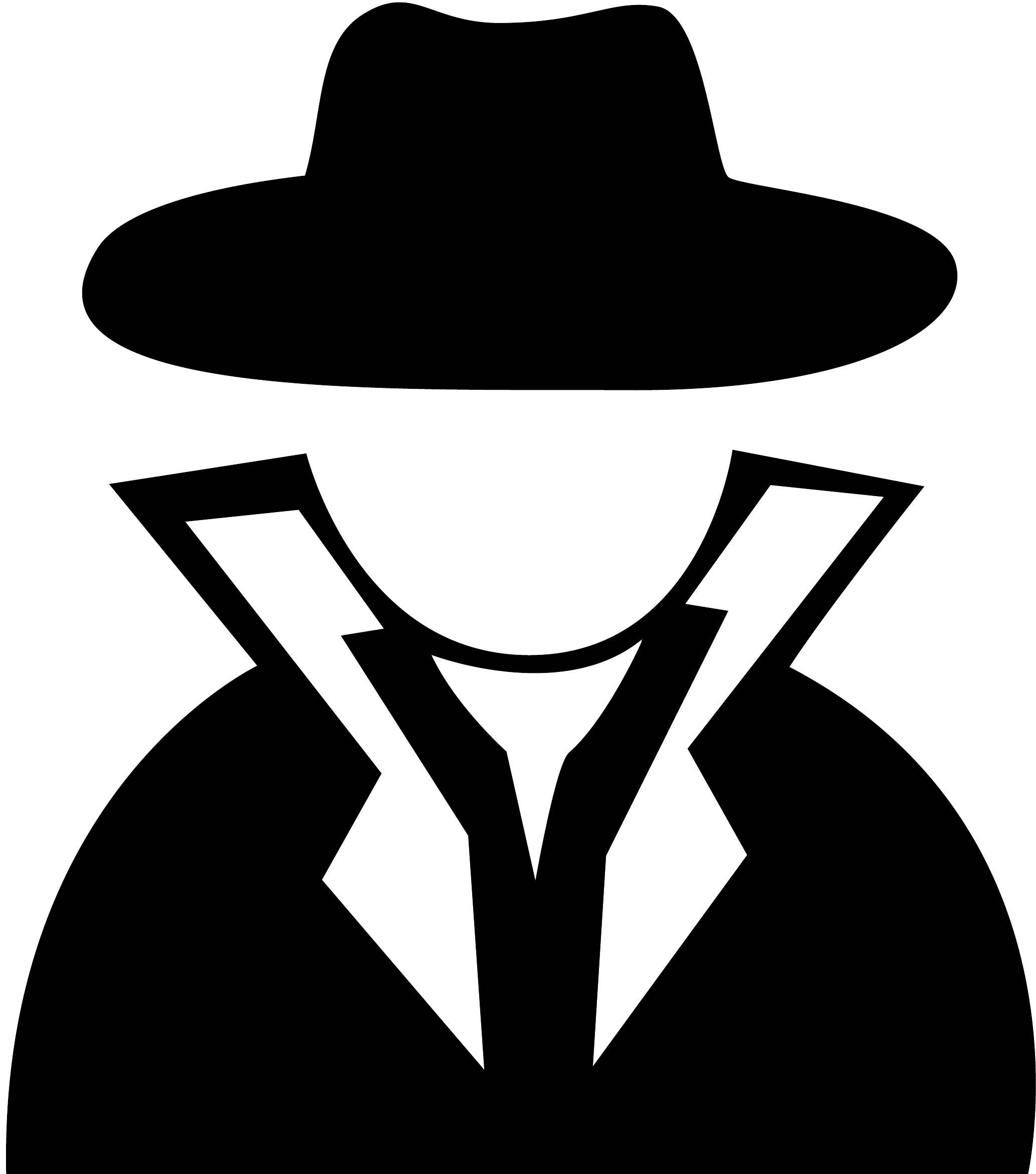}};
    \node[imgnode, label={below:\(W1\)}] (w1) at (w1coord){\includegraphics[width=0.75cm]{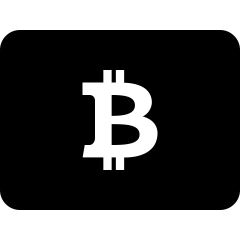}};
    \node[imgnode, label=below:{\(W2\)}] (w2) at (w2coord){\includegraphics[width=0.75cm]{figures/icons/walletFree.png}};

    \draw [pen colour={black},
    decorate,
    decoration = {calligraphic brace,
      raise=5pt,
      amplitude=5pt,
      aspect=0.5}] ($(dudebuy)+(0,1)$ )--  ($(w2)+(0,2)$)
    node[pos=0.5,above=10pt,black, align = center]{Suspicion through\\Address Control};

    \node[imgnode, label={below:\(W4\)}] (w4) at (w4coord){\includegraphics[width=0.75cm]{figures/icons/walletFree.png}};
    ;

    \node[imgnode, label={below:Mixer}] (mixer) at (mixercoord)
    {\includegraphics[width=1cm]{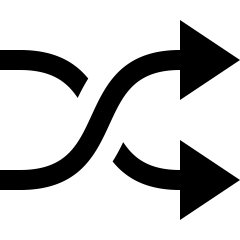}};

    \node[imgnode, label={below:BPPC}] (bppc) at (bpcccoord){\includegraphics[width=0.75cm]{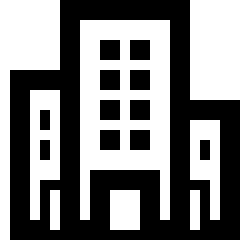}};
    \draw [pen colour={black},
    decorate,
    decoration = {calligraphic brace,
      raise=5pt,
      amplitude=5pt,
      aspect=0.5}] ($(w1)+(0,1.2)$ )--  ($(bppc)+(0,1)$)
    node[pos=0.5,above=10pt,black, align = center]{Cluster\\from Software};

    \node[imgnode, label={below:Game Company}] (shopping) at (shoppingcoord){\includegraphics[width=0.75cm]{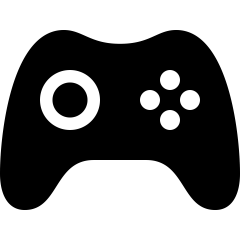}};

    \draw [pen colour={black},
    decorate,
    decoration = {calligraphic brace,
      raise=5pt,
      amplitude=5pt,
      aspect=0.5}] ($(w4)+(0,-1)$) -- ($(w2)+(0,-1.75)$)
    node[pos=0.5,below=10pt,black, align = center]{Argument\\from Sign};

    \node[imgnode, label={below:E-Mail}] (email) at (emailcoord){\includegraphics[width=0.75cm]{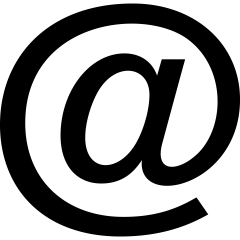}};

    \node[below = 3em of shopping, imgnode, label={below:Defendant X.}] (x) {\includegraphics[width=0.95cm]{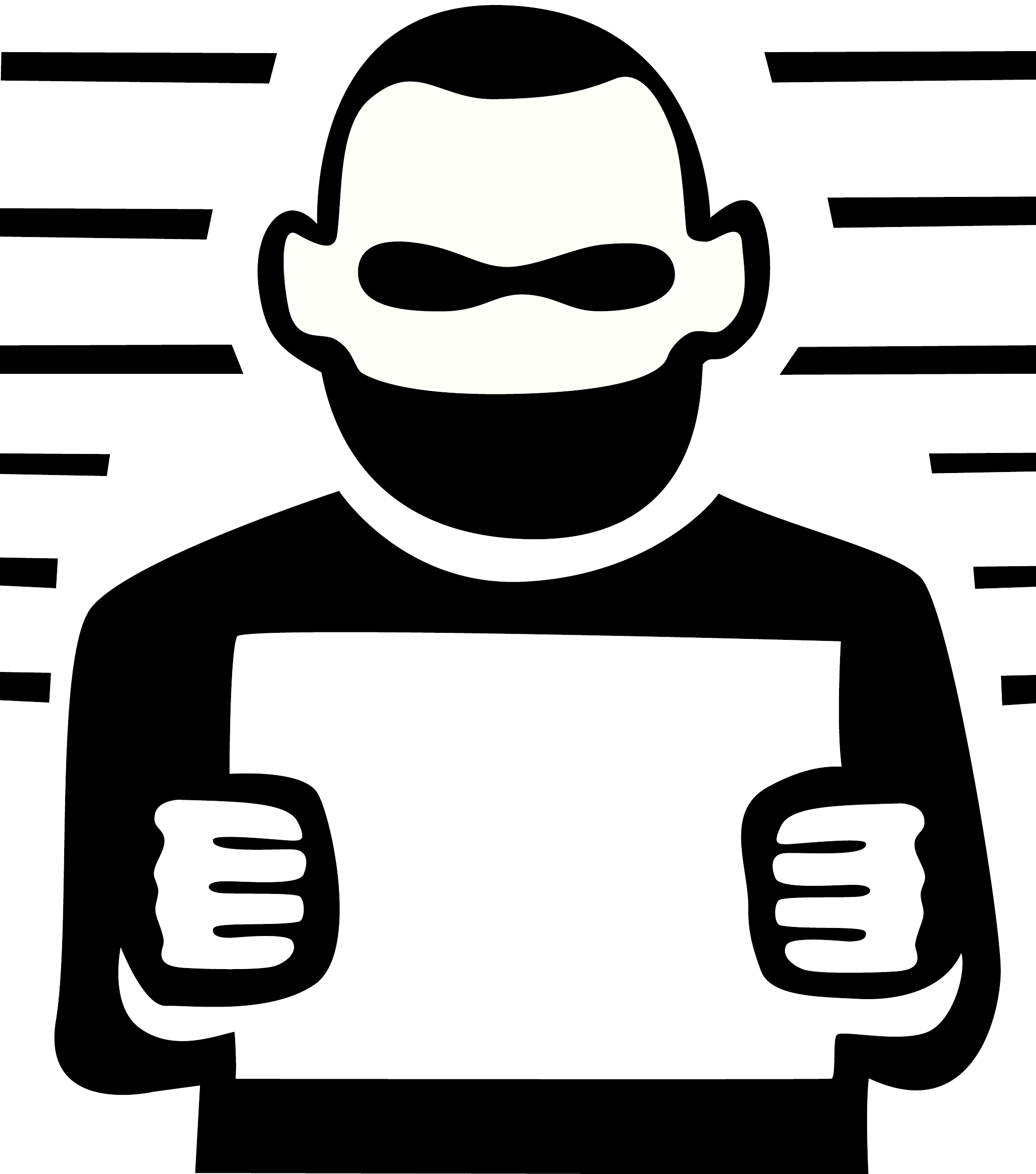}};

    \draw [pen colour={black},
    decorate,
    decoration = {calligraphic brace,
      raise=5pt,
      amplitude=5pt,
      aspect=0.5}] ($(shopping)+(0,2)$ )--  ($(email)+(0,1)$)
    node[pos=0.5,above=10pt,black, align = center]{Argument from\\Position to Know};

    \draw [black,
    decorate,
    decoration = {brace,
      raise=5pt,
      amplitude=5pt,
      aspect=0.5}] ($(email)+(0.8,-.2)$) -- ($(shopping)+(1.3 ,-.8)$)
    node[midway, rotate=58.2,yshift=-20pt,black, align = left]{Argument\\from Sign};

    \draw [pen colour={black},
    decorate,
    decoration = {calligraphic brace,
      raise=5pt,
      amplitude=5pt,
      aspect=0.5}] ($(email)+(0,-3.8)$) -- ($(bppc)+(0,-3.8)$ )
    node[pos=0.5,below=10pt,black, align = center]{Argument from\\Position to Know};

    \draw[bend right] (x) edge ($(email)+(0,-0.7)$);
    \draw[bend right] ($(bppc)+(0,-0.7)$) edge (x);

    \draw[] (dudebuy) edge (w2);

    \draw[] ($(dudebuy)+(0.07,-0.7)$) edge ($(hansa)+(0.07,0.4)$);
    \draw[-Latex] (hansa) edge (w2);
    \draw[-Latex] (w2) edge (w1);
    \draw[-Latex] (w2) edge (w4);
    \draw[-Latex] (w1) edge (mixer);
    \draw[-Latex] (w4) edge (mixer);
    \draw[-Latex] (mixer) edge (bppc);
    \draw[-Latex] (bppc) edge (email);
    \draw[-Latex] (bppc) edge (shopping);

    \draw[-Latex] (shopping) edge (email);
  \end{tikzpicture}
  \caption{Application of the proposed argumentation schemes to assess the identification of the administrator of the darknet marketplace called Wall Street Market\label{fig:wsm_illustration}}
\end{figure}
 The
publicly available criminal complaint states that the \ac{USPS} employed proprietary software of an undisclosed company to conduct its blockchain analyses~\cite{complaint2019wsm}.
Furthermore, neither the exact methods employed during the analyses nor the involved Bitcoin addresses were specified. Instead, the final results~--~meaning actual investigative findings in the form of off-chain information~-- were presented on their own.
To prove the correctness, it is merely stated that the software was found to be reliable based on numerous unrelated investigations~\cite{complaint2019wsm}. 
This might either suggest the software was utilised as a black box or that the details were (intentionally) not published and kept secret to protect the technical means for tactical reasons. 
This argumentation might be insufficient to convince legal decision makers of the rightfulness of the findings.
Thus, we infer from the criminal complaint which analysis methods the software might have employed and then apply our argumentation schemes to argue the findings.  

The blockchain analyses of the \ac{USPS} constituted the initial lead that enabled the
involved law enforcement agencies
to identify \enquote{TheOne} -- who acted as one of the administrators of the
platform~\cite{complaint2019wsm}. \enquote{TheOne} is believed to be
\enquote{X.},\footnote{The defendant's name has been anonymized by the
  authors.} one of the three defendants, mainly based on the following two findings:
%

First, the investigators could establish a link between the administrator \enquote{TheOne} from
\ac{WSM} and the user \enquote{dudebuy} from \emph{Hansa Market} by analysing data seized
from both platforms. They found that \enquote{TheOne} used the same PGP public key
as
\enquote{dudebuy} did at the previously operated and
meanwhile seized darknet marketplace Hansa Market. As a PGP key pair is a highly
individual piece of data used to prove one's identity and encrypt communications, it has to
be inferred that those two monikers belong to the same real-world entity. As
\enquote{dudebuy} used a wallet $W2$ as his \emph{refund wallet} on Hansa Market, the
investigators found an entry point to perform financial investigations concerning this
perpetrator seeming to operate now as \enquote{TheOne}.

Here, the investigators could establish suspicion using the \textit{\nameref{fig:suspicion}} scheme
and infer that the owner of wallet \(W2\) seems to be the targeted
administrator of the ongoing investigations regarding \ac{WSM}.
This conclusion could be assessed by the evaluation of the \aclp{CQ} of the scheme.
\Cref{enum:arg-suspicion:q1} -- regarding circumstantial evidence indicating address control -- leads to a high degree of
confidence, as the investigators resorted to seized user data, including an identical
PGP public key. While \cref{enum:arg-suspicion:q2} (address control by somebody else) does not seem to be of relevance to the investigators
at this point in time, \cref{enum:arg-suspicion:q3} (nature of the connection to the offence)  
reveals at least an indirect involvement of the address in the offence
in question.

Second, being confident that the owner of wallet $W2$ is the target, the \ac{USPS} revealed that other wallets that appeared in the investigations, namely wallets $W1$ and $W4$, were funded by transactions originating from wallet $W2$.
As this analysis step is basically a rather typical payment flow analysis, which is also employed in traditional money laundering investigations concerning fiat currencies, it is dispensable to assess it with a
newly formulated argumentation scheme. For example, \emph{Argument from Sign} or \emph{Argument from Abductive Inference} would be a suitable fit here~\cite{walton_reed_macagno_2008}. Those newly uncovered wallets, in turn, were identified to be the true origin of several payments to various services, which were conducted via a \ac{BPPC}. Prior to these payments, the corresponding funds were supposedly mixed via a commercial mixing service, whose flow of transactions could be \enquote{de-mixed} by the \ac{USPS}' analysts~\cite{complaint2019wsm}.

Given the fact that no further information regarding the de-mixing is
presented in the criminal complaint, we deliberately assume that some
sort of software established the link so that the
\textit{\nameref{fig:blackbox}} scheme should be employed to be able to judge
the evidential value of this result. The scheme revolves around the
mechanism for link establishment (\cref{enum:arg-blackbox:q1}),
the reliability of the tool itself~(\cref{enum:arg-blackbox:q2}),
human comprehensibility~(\cref{enum:arg-blackbox:q3}) and additional evidence available
(\cref{enum:arg-blackbox:q4,enum:arg-blackbox:q5}).
Here, the most important critical question to pose might be
\cref{enum:arg-blackbox:q3}, i.e.\ whether the link could be
established by comprehensible reasoning of a human analyst. 
As the following requests for the disclosure of customer data were based on this link, it must be considered crucial evidence in this early phase of the investigation.
In the course of using \cref{enum:arg-blackbox:q3}, a human analyst might establish that the link was a result of the multi-input heuristic. 
As the multi-input heuristic results in false positives when applied to CoinJoin transactions, it is crucial to challenge whether the involved transactions could be CoinJoin transactions -- via \cref{enum:arg-mih:q1} of the \textit{\nameref{fig:multi-input}} scheme.
By this example, the practical relevance of our argumentation schemes becomes particularly apparent. 
Without the schemes, the argumentation would be limited to whether the analysis software was reliable in the past but not whether false positives were actually excluded in the specific case.

By obtaining user records from the \ac{BPPC} regarding the payment from wallet $W1$, investigators uncovered an e-mail address, which could be linked to the aforementioned defendant, as it was actually used alongside his real-world identity 'X'. In addition to that, 
they uncovered that wallet $W4$ served as the suspected source for payments for two accounts at a video gaming company, 
which were also linked to the suspect, as the records obtained by a subpoena suggest. 
Furthermore, a second link could be established from
another wallet $W5$, in a similar manner, which is considered to be used to pay for a third account linked to the suspect at the gaming company in a similar manner. 
Wallet $W5$ was found to be funded by a different wallet that could also be associated with \ac{WSM}'s administrators at a later point in time. While this correlation accumulates reliability, each respective 
request for the disclosure of customer data might be assessed by employing the \emph{Argument from Position to Know} scheme~\cite{walton_reed_macagno_2008}.

In summary, the \ac{USPS}'s blockchain analyses included the following broader steps:
\emph{identification of wallets}, \emph{detection of payments between wallets},
\emph{de-mixing} and the \emph{association of wallets with off-chain information} mainly
from other darknet marketplaces as well as service providers.
While the investigators later found various pieces of evidence in the course of the
following investigative actions, these steps were central for the case in order to find a
starting point for targeted investigations. We showed that their reliability could be effectively assessed by the utilisation of our argumentation schemes.

\section{Conclusion}\label{sec:discussion}

After having demonstrated the usage of several argumentation schemes for
blockchain-based investigations, we conclude by presenting use cases in which
the schemes will be especially beneficial and by pointing out directions for future work.

As our argumentation schemes allow reasoning about the findings of blockchain-based investigations, we see potential use cases wherever such findings have to be communicated to and assessed by persons involved in respective criminal proceedings.
By utilising the schemes, an analyst can clearly articulate the employed heuristics, their individual strengths, and potential weaknesses.
This increases the comprehensibility of such analyses and court proceedings for the decision makers, and also eases the documentation for later verification by an expert witness.
Given the high requirements regarding the explainability of legal proceedings, this task cannot be achieved by software in an automated manner yet. 
Therefore, we intend to support them with our argumentation schemes.
Nevertheless, our considerations can be prospectively integrated into deanonymization software to increase its explainability.
Clear articulation is key to
determining the quality of blockchain-based findings, especially if they are not or only weakly supported by other evidence.
On the one hand, applying an argumentation scheme and utilising its critical questions enables law enforcement agencies and the preliminary judge
to reason about the eventual perpetration of the identified person and therefore establish a certain degree of suspicion to justify further investigative measures. 
On the other hand, the rights of suspects can be protected
by ensuring that the results obtained from blockchain investigations are of quality, can be understood, independently checked for plausibility by the parties to the proceedings, and are actually able to establish the relevant suspicion required by law.

As a result, we consider the application of argumentation schemes in the context of blockchain-based investigations a supportive mechanism for making sense of the
intangible crime scene and highly abstract commission of cybercriminal offences.
Our schemes can be a helpful tool for investigators and prosecutors that strive to identify perpetrators, as well as for legal decision makers to answer the question of guilt.
Finally, the schemes are a step forward in the direction of harmonising the effectiveness and explainability of high-tech investigations.

Extending this work can be done in multiple directions.
Further schemes for other blockchain analysis heuristics or other cybercriminal investigations could be created, as indicated already in~\cref{sec:our-argum-schemes}. 
In addition to that, the \aclp{CQ} of our schemes could be refined to comprise more specific sub-questions as done for \emph{Argument from Expert Opinion} in~\textcite{walton_reed_macagno_2008} to capture more expert knowledge.

\subsubsection*{Acknowledgements}
This work was supported by DFG (German Research Foundation) as part of the Research and Training Group 2475 ``Cybercrime and Forensic Computing'' (grant number~393541319/GRK2475/1-2019). 
Merlin Humml was also supported by DFG project RAND (grant number~377333057).
The authors also wish to thank Marie-Helen Maras for fruitful discussions.

\printbibliography

\end{document}